\documentclass[10pt,conference]{IEEEtran}
\ifCLASSINFOpdf
  \usepackage[pdftex]{graphicx}
  \graphicspath{{./img/}}
  \DeclareGraphicsExtensions{.pdf,.png}
\else
\fi

\usepackage{xcolor}

\usepackage{amsmath,amssymb,amsthm,mathtools}
\usepackage{algorithm}
\usepackage{algorithmic}
\usepackage{enumitem}
\usepackage{booktabs}
\usepackage{bbm}

\usepackage{subcaption} 

\usepackage{enumitem}
\usepackage{scalerel}

\AtBeginDocument{%
  \addtolength{\topmargin}{0.025in}%
  \addtolength{\textheight}{-0.050in}%
}

\allowdisplaybreaks

\newcommand{\ba}{\begin{array}}
	\newcommand{\ea}{\end{array}}

\makeatletter
\newcommand\fs@betterruled{%
  \def\@fs@cfont{\bfseries}\let\@fs@capt\floatc@ruled
  \def\@fs@pre{\vspace*{8pt}\hrule height.8pt depth0pt \kern2pt}%
  \def\@fs@post{\kern2pt\hrule\relax}%
  \def\@fs@mid{\kern2pt\hrule\kern2pt}%
  \let\@fs@iftopcapt\iftrue}
\floatstyle{betterruled}
\restylefloat{algorithm}
\makeatother

\begin{document}


\title{FedQoS: Federated QoS-Risk Learning for Heterogeneous Indoor-Outdoor Access Selection}

\author{
    \IEEEauthorblockN{Nguyen Van Thieu, Ti Ti Nguyen, Ons Aouedi, Zerihun Huruy, Vu Nguyen Ha, and Symeon Chatzinotas}
    \IEEEauthorblockA{\textit{Interdisciplinary Centre for Security, Reliability and Trust (SnT), University of Luxembourg, Luxembourg}\\
    \{vanthieu.nguyen, titi.nguyen, ons.aouedi, huruy.zerihun,
vu-nguyen.ha, symeon.chatzinotas\}@uni.lu} 
}

\maketitle

\begingroup
\renewcommand\thefootnote{}
\footnotetext{%
\copyright~2026 IEEE. Personal use of this material is permitted.
Permission from IEEE must be obtained for all other uses, in any
current or future media, including reprinting/republishing this
material for advertising or promotional purposes, creating new
collective works, for resale or redistribution to servers or lists,
or reuse of any copyrighted component of this work in other works.%
}
\addtocounter{footnote}{-1}
\endgroup

\begin{abstract}
Reliable access selection in dynamic and heterogeneous indoor-outdoor environments is challenging because instantaneous radio measurements alone cannot capture future QoS degradation caused by mobility, blockage, traffic load, and resource competition. This paper proposes FedQoS, a federated QoS-risk learning framework for predicting the future reliability of candidate access links and supporting access-node selection without centralizing user-level network data. In FedQoS, each access node locally learns from its observed network logs, including radio, traffic, load, and service-context features, while a global QoS-risk predictor is trained through federated aggregation. The learned model estimates the probability of QoS failure for each candidate link, and the controller uses these risk scores to select reliable access nodes under dynamic network conditions. To evaluate the framework, we construct physics-based synthetic indoor-outdoor wireless datasets using the Sionna framework, covering normal traffic, mobility, event-driven congestion, and non-IID client observations. 
Simulation results show that learning-based access selection substantially reduces the QoS-failure rate compared with signal-based and historical-QoS heuristic methods. 
FedQoS achieves near-centralized predictive performance and provides
clear reliability gains under mild non-IID data while remaining
competitive under the more challenging severe non-IID condition.
These results demonstrate the potential of federated QoS-risk learning for reliable, data-local access selection in dynamic wireless environments.

\end{abstract}


%
\IEEEpeerreviewmaketitle

\section{Introduction}
\label{sec:introduction}

Heterogeneous wireless access networks are becoming a key component of future indoor-outdoor connectivity. In dense environments such as enterprise buildings, public venues, transportation hubs, and university campuses, users move across rooms, corridors, building entrances, outdoor walkways, and temporarily crowded areas, while their candidate serving nodes span indoor access points (APs), terrestrial base stations (BSs), and static or quasi-static aerial nodes. Deciding which node should serve each user at a given time, the user association or access selection problem, directly affects throughput, load balancing, delay, handover behavior, and quality of service~\cite{liu2016user}.

Conventional  network selection mechanisms  are based on radio-layer metrics such as received signal strength, received power, or SINR~\cite{kim2023distributed}. While these signal-oriented criteria are straightforward and interpretable, they frequently fall short in heterogeneous networks. For example, a user attached to the node with the strongest signal may still experience poor service if that node is heavily loaded, whereas an alternative node with a weaker signal may deliver higher long-term throughput or reduced latency. This limitation is well recognized in the  literature; max-SINR or max-power association can cause severe load imbalance~\cite{ye2013user}, motivating biasing and load balancing schemes~\cite{liu2016user}, while QoS and delay-aware methods further incorporate attainable rate, load, required resources, or queueing delay into the association decision~\cite{zhou2014qos, kong2017delay}. These studies establish that the association should not be driven by instantaneous radio strength alone.

Learning-based methods have since been investigated for association and radio resource management. Deep reinforcement learning (DRL) and multi-agent learning are applied when explicit optimization is difficult due to non-convexity, dynamic, or incomplete data. On the other hand, federated learning (FL) has been introduced to reduce raw-data sharing in wireless control, e.g., federated DRL for user access control in Open RAN and for distributed control of NextG networks~\cite{cao2021user, tehrani2021federated}. Despite this progress, 
most learning-based association works formulate the problem as direct
control or RL and learn a policy from a reward signal~\cite{zhao2019deep}. Such policies can be difficult to interpret and tune because their behavior depends strongly on the reward design.
Moreover, existing federated wireless-control frameworks~\cite{lim2020federated} mainly target policy learning, resource allocation, or user-side learning, rather than access node-side supervised QoS-risk prediction from local logs~\cite{bega2019deepcog}. Yet in practical deployments, each AP, BS, or aerial node naturally observes local user-candidate measurements, load states, service demands, handover context, and realized QoS outcome, exactly the signals needed to learn whether a choice will lead to a QoS violation over a short future horizon~\cite{mollel2021survey}.

This paper proposes FedQoS, a federated QoS-risk learning framework for heterogeneous indoor-outdoor access selection. Instead of directly learning an association action, FedQoS learns a supervised probabilistic model that predicts the future QoS-failure risk of each user-candidate pair, i.e., the probability that a candidate AP, BS, or static aerial node will violate the user's QoS requirement within a prediction window.
The QoS-risk predictor incorporates access-node load through its input
features, while the access controller combines the resulting risk score
with an explicit handover cost to select the serving node.
The federated design follows the distributed nature of access-node logs, where each node trains on its own observations and shares only model updates. Because these clients are naturally non-IID (differing in user density, mobility, load, and QoS-failure rates), FedQoS couples stabilized local training with a QoS-aware aggregation weight that reflects both local data volume and local QoS-failure exposure, so that nodes observing QoS-critical conditions contribute more without discarding the statistical reliability of larger clients.

The main contributions of this paper are summarized as follows:
\begin{itemize}
    \item Formulate heterogeneous indoor-outdoor access selection as a federated supervised QoS-risk prediction problem over user-candidate access pairs.
    \item Propose FedQoS combining stabilized local training with a QoS-aware aggregation rule to handle heterogeneous, non-IID observations from APs/BSs/aerial nodes.
    \item Develop a reliability-oriented access-selection rule that converts predicted QoS-failure probabilities into access decisions, with access-node load captured by the learned risk model and handover cost handled explicitly by the controller.
    \item Construct realistic dynamic datasets using ray-tracing and system-level wireless simulation and evaluate FedQoS against centralized, local-only, FedAvg, and FedProx learning and representative non-learning baselines.
\end{itemize}

\section{System Model and Proposed FedQoS Algorithm}
\label{sec:system_fedqos}

\subsection{Heterogeneous Indoor-Outdoor Access Setting}
\label{subsec:access_setting}

\begin{figure*}[!t]
\includegraphics[width=\linewidth]{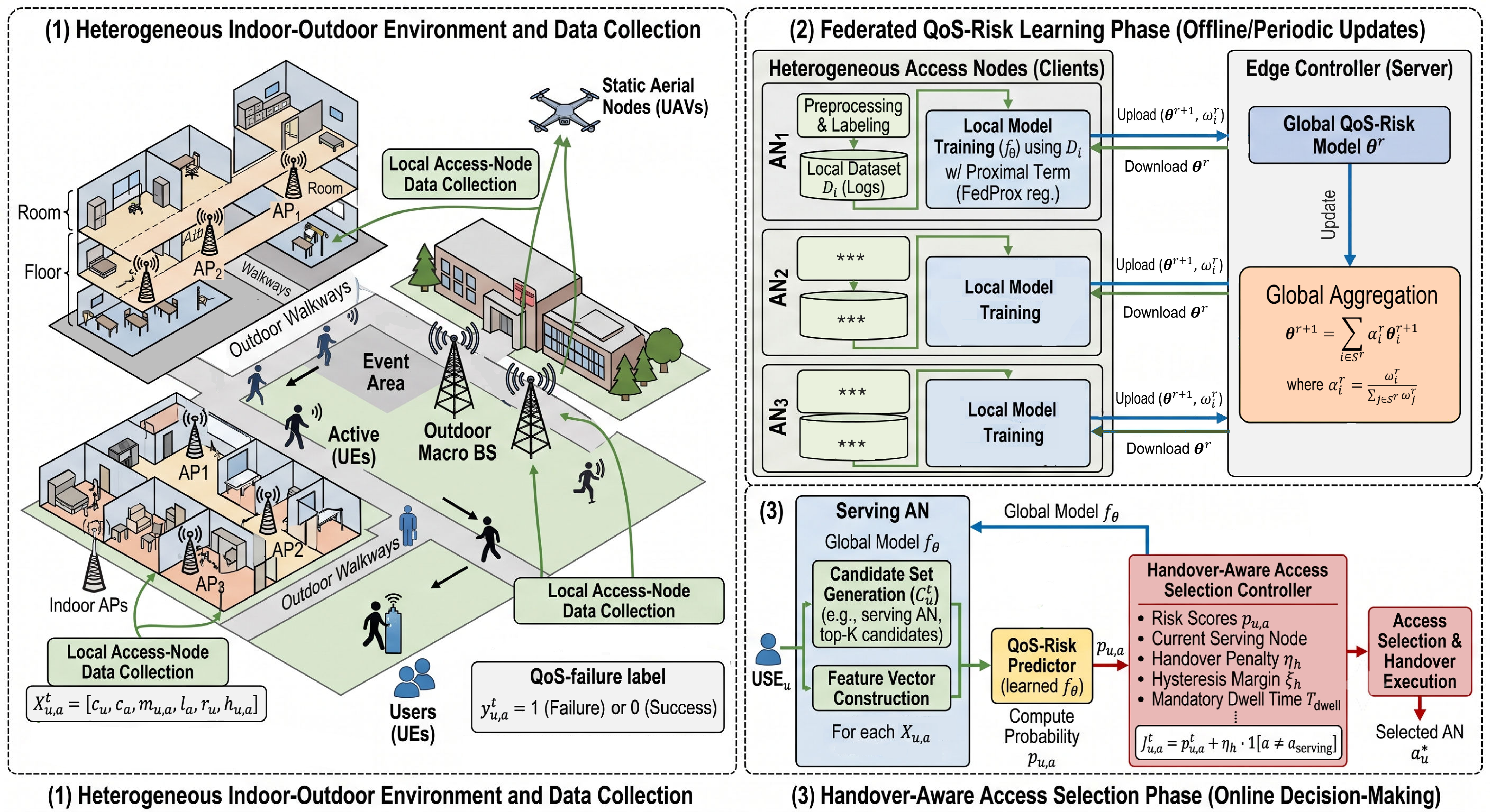}
\caption{An example of an indoor-outdoor access network and the
FedQoS system.}
\label{fig:SAGIN}
\end{figure*}

We consider a heterogeneous indoor-outdoor access network composed of indoor APs, terrestrial BSs, and static aerial access nodes (ANs) serving a varying number of users within a campus area, as illustrated
in Fig.~\ref{fig:SAGIN}. 
The ANs are considered to provide a flexible 
connection to the users. In the following, it is assumed that their positions are fixed during the considered decision process of $T$ time slots. It is worth noting that trajectory and placement optimization are out of the scope of this work.

Let $\mathcal{A}$ be the set of all access nodes, and denote by $\mathcal{U}^{t}$ the set of active users
in slot $t$, where $t \in  \{0,1,\ldots,T-1\}$. 
This work aims to develop an FL-based access controller to instruct users to select the most suitable access node for their next communication. In particular, in the slot $t$,
the controller can select an access node $a_u^t\in\mathcal{A}$ from the candidate set $\mathcal{C}_{u}^{t} \subseteq \mathcal{A}$ that contains a limited number of feasible service nodes selected from the measurement report. In the implementation, the candidate set is formed from the current serving node, the top-$K$ candidates according to the logging association score, and occasional probing candidates. To maintain the users' QoS, the access node should be selected for each active user so that the corresponding transmission rate for that connection is not less than $R_u^{\min}$.

To build such a controller, in this system, each access node acts as a federated client where its local data set is denoted by $\mathcal{D}_i$. A user--candidate sample observed while user $u$ is served by node $a_u^t$ is stored at the serving node. Therefore, the observer client for that sample is $o_u^t = a_u^t$,
and the sample belongs to $\mathcal{D}_{o_u^t}$. This construction matches practical operation, where the access node can collect its users' measurements, service context, recent history, and candidate reports.

The objective of the access controller is to select, for each user, an access node that is likely to satisfy the user's quality-of-service (QoS) requirement over a short future horizon. Unlike conventional access selection based only on the strongest received signal, the considered environment is affected by indoor blockages, floor transitions, wall penetration loss, dynamic user density, access-node load, and heterogeneous AP/BS/AN coverage. Therefore, the access node with the strongest instantaneous signal is not necessarily the one that provides the lowest QoS-failure risk.

\subsection{User--Candidate Feature Vector}
\label{subsec:feature_vector}

For user $u$ and access node $a$, the local controller constructs a feature vector for slot $t$ as
\begin{equation}
    \mathbf{x}_{u,a}^t =
    \big[
        \mathbf{c}_u^t,\,
        \mathbf{c}_a,\,
        \mathbf{m}_{u,a}^t,\,
        \mathbf{l}_a^t,\,
        \mathbf{r}_u^t,\,
        \mathbf{h}_{u,a}^t
    \big],
    \label{eq:feature_vector}
\end{equation}
where \(\mathbf{c}_u^t\) denotes user-side context such as coarse indoor/outdoor zone, floor index, mobility state, and radio-positioning features; \(\mathbf{c}_a\) denotes access-node attributes such as node type, carrier, bandwidth, transmit power, antenna configuration, and height; \(\mathbf{m}_{u,a}^t\) contains instantaneous or recently averaged link measurements such as reference signal received power (RSRP), received signal strength indicator (RSSI), signal-to-interference-plus-noise ratio (SINR), channel quality indicator (CQI), block error rate (BLER), or achievable-rate indicators; \(\mathbf{l}_a^t\) contains access-node load information such as active-user count, physical resource block (PRB) utilization; \(\mathbf{r}_u^t\) represents the user service requirement, including \(R_u^{\min}\); and \(\mathbf{h}_{u,a}^t\) contains handover-related indicators such as whether \(a=a_u^{t-1}\), dwell time, and recent handover history.

The proposed framework does not require raw user logs to be uploaded to a central server. Each access node stores the logs observed for users served or probed by that node. The central controller receives only model parameters or model updates during federated training.

\subsection{Future QoS-Failure Label}
\label{subsec:qos_label}

The learning target is the future QoS outcome of a candidate access decision. Let $\Delta$ be the prediction horizon. If user $u$ is associated with candidate access node $a$ at slot $t$, let $\bar{R}_{u,a}^{t,\Delta}$ and $\bar{\epsilon}_{u,a}^{t,\Delta}$
denote the average throughput and BLER over the future window $[t,t+\Delta]$. Given the min-rate requirement $R_u^{\min}$ and max tolerable BLER $\epsilon_{\max}$, the QoS-failure label is defined as
\begin{equation}
    y_{u,a}^{t}
    =
    \mathbf{1}_
    {
    \left(\bar{R}_{u,a}^{t,\Delta}<R_u^{\min}\right)
    \;\; \mathrm{or} \;\;
    \left(\bar{\epsilon}_{u,a}^{t,\Delta}>\epsilon_{\max}\right)
    }.
    \label{eq:qos_failure_label}
\end{equation}
Thus, $y_{u,a}^{t}=1$ means that candidate node $a$ is expected to violate the QoS requirement of user $u$ over the prediction horizon, whereas $y_{u,a}^{t}=0$ means that the candidate is expected to satisfy the requirement.

The learning target is not the instantaneous SINR alone. SINR is treated as one of the explanatory features. The label in \eqref{eq:qos_failure_label} captures the combined effect of radio quality, resource contention, service demand, and mobility over a short future horizon. This formulation is useful because two candidate links with similar SINR can have different QoS outcomes due to different load, scheduling, or handover conditions.

\subsection{QoS-Risk Learning}
\label{subsec:fl_learning}

Let \(\mathcal{D}_i=\{(\mathbf{x}_{n},y_{n})\}_{n=1}^{N_i}\) denote the local dataset stored at access node \(i\in\mathcal{A}\), where \(N_i=|\mathcal{D}_i|\). The datasets are naturally non-IID because access nodes observe different locations, building structures, user densities, channel conditions, service mixes, and mobility patterns. For example, indoor APs mainly observe room and corridor users, whereas BSs and UAVs observe more outdoor or overloaded-event users.

The QoS-risk model is denoted by $f_{\boldsymbol{\theta}}$, where $\boldsymbol{\theta}$ is the model parameter vector. Given the input feature vector $\mathbf{x}_{u,a}^{t}$, the model outputs
\begin{equation}
    p_{u,a}^{t}
    =
    f_{\boldsymbol{\theta}}
    \left(
        \mathbf{x}_{u,a}^{t}
    \right)
    \in [0,1],
    \label{eq:qos_risk_prediction}
\end{equation}
where $p_{u,a}^{t}$ estimates the probability that candidate $a$ will fail the QoS requirement of user $u$ over the prediction horizon:
\begin{equation}
    p_{u,a}^{t}
    \approx
    \Pr
    \left(
        y_{u,a}^{t}=1
        \mid
        \mathbf{x}_{u,a}^{t}
    \right).
\end{equation}

The local empirical loss at client $i$ is
\begin{equation}
    F_i(\boldsymbol{\theta})
    =
    \frac{1}{|\mathcal{D}_i|}
    \sum_{(\mathbf{x},y)\in\mathcal{D}_i}
    \ell
    \left(
        f_{\boldsymbol{\theta}}(\mathbf{x}),y
    \right),
    \label{eq:local_loss}
\end{equation}
where $\ell(\cdot,\cdot)$ is the binary cross-entropy loss. Then the conventional centralized solution would minimize
\begin{equation}
    \min_{\boldsymbol{\theta}}
    \sum_{i\in\mathcal{A}}
    \frac{N_i}{N}
    F_i(\boldsymbol{\theta}),
    \qquad
    N=\sum_{i\in\mathcal{A}}N_i,
    \label{eq:centralized_obj}
\end{equation}
which requires collecting all local access logs at a central server. FedQoS instead solves this problem federatively, where each access node keeps \(\mathcal{D}_i\) locally and exchanges only model parameters with an edge controller.

\subsection{FedQoS Local Update}
\label{subsec:fedqos_local}

At communication round \(r\), the edge controller broadcasts the current global model \(\boldsymbol{\theta}^{r}\) to a subset of participating access nodes \(\mathcal{S}^r\subseteq\mathcal{A}\). Each selected client \(i\in\mathcal{S}^r\) initializes its local model by
$\boldsymbol{\theta}_{i,0}^{r} = \boldsymbol{\theta}^{r}$.
FedQoS builds its local training step on a proximal objective \cite{li2020federated} to stabilize learning under heterogeneous local data. Specifically, client \(i\) minimizes
\begin{equation}
    \widetilde{F}_i(\boldsymbol{\theta};\boldsymbol{\theta}^{r})
    =
    F_i(\boldsymbol{\theta})
    +
    \frac{\mu}{2}
    \left\|
        \boldsymbol{\theta}-\boldsymbol{\theta}^{r}
    \right\|_2^2,
    \label{eq:fedqos_prox_obj}
\end{equation}
where \(\mu\geq 0\) penalizes the deviation of the local model from
the current global model.
After \(E\) local epochs, client \(i\) obtains \(\boldsymbol{\theta}_{i}^{r+1}\) and uploads it to the edge controller.

\subsection{FedQoS Aggregation}
\label{subsec:qos_aggregation}

A standard sample-size weighted aggregation can be dominated by high-traffic access nodes, such as a BS serving many outdoor users. However, for QoS-risk learning, clients that observe many rare failure events are also important. FedQoS therefore uses a QoS-aware aggregation rule that balances sample size and local risk.
We define the local QoS-failure rate as 
$\rho_i^r = \frac{1}{N_i} \sum_{(\mathbf{x},y)\in\mathcal{D}_i} y$ 
and the unnormalized FedQoS aggregation weight as follows:
\begin{equation}
    \omega_i^r
    =
    \left(N_i\right)^{q}
    \left(1+\lambda \rho_i^r\right),
    \qquad i\in\mathcal{S}^r.
    \label{eq:fedqos_weight}
\end{equation}
where \(q\in[0,1]\) controls the dependence on client dataset size and \(\lambda\geq 0\) controls the emphasis on clients that observe more QoS failures. Setting \(q=1\) and \(\lambda=0\) recovers standard sample-size weighted aggregation. Smaller values of \(q\) reduce the domination of very large clients, while positive \(\lambda\) increases the influence of high-risk clients. 
Then the normalized aggregation coefficient can be calculated as
\begin{equation}
    \alpha_i^r
    =
    \frac{\omega_i^r}
    {\sum_{j\in\mathcal{S}^r}\omega_j^r},
    \qquad i\in\mathcal{S}^r.
    \label{eq:fedqos_alpha}
\end{equation}
The global model is updated as
\begin{equation}
    \boldsymbol{\theta}^{r+1}
    =
    \sum_{i\in\mathcal{S}^r}
    \alpha_i^r
    \boldsymbol{\theta}_{i}^{r+1}.
    \label{eq:fedqos_update}
\end{equation}
This aggregation rule encourages the global model to preserve information from high-risk access regions without ignoring the statistical reliability of larger local datasets.

\begin{algorithm}[!t]
\caption{FedQoS: Federated QoS-Risk Learning and Handover-Aware Access Selection}
\label{alg:fedqos}
\begin{algorithmic}[1]
\REQUIRE Access nodes \(\mathcal{A}\), local datasets
\(\{\mathcal{D}_i\}_{i\in\mathcal{A}}\), rounds \(R\), epochs \(E\), coefficient \(\mu\), \(q\), \(\lambda\), \(\eta_h\), $\epsilon_{max}$, $\xi_h$, and $T_{dwell}$.
\ENSURE Global QoS-risk model \(f_{\boldsymbol{\theta}^{R}}\).
\STATE Initialize global model parameters \(\boldsymbol{\theta}^{0}\).
\FOR{round \(r=0,1,\ldots,R-1\)}
    \STATE Edge server selects participating clients
    \(\mathcal{S}^r\subseteq\mathcal{A}\).
    \STATE Broadcast \(\boldsymbol{\theta}^{r}\) to all clients
    \(i\in\mathcal{S}^r\).
    \FOR{each client \(i\in\mathcal{S}^r\) in parallel}
        \STATE Set local model \(\boldsymbol{\theta}_{i,0}^{r}=\boldsymbol{\theta}^{r}\) and train it using \eqref{eq:fedqos_prox_obj}.
        \STATE Obtain updated local model
        \(\boldsymbol{\theta}_{i}^{r+1}\).
        \STATE Compute aggregation weight $\omega_i^r$ by \eqref{eq:fedqos_weight}
        \STATE Upload \(\boldsymbol{\theta}_{i}^{r+1}\) and \(\omega_i^r\) to the controller
    \ENDFOR
    \STATE Compute aggregation coefficients \(\{\alpha_i^r\}_{i\in\mathcal{S}^r}\) by \eqref{eq:fedqos_alpha}
    \STATE Update the global model using \eqref{eq:fedqos_update}.
\ENDFOR
\STATE Deploy \(f_{\boldsymbol{\theta}^{R}}\) to all access nodes.
\FOR{each access-decision slot \(t\)}
    \FOR{each active user \(u\in\mathcal{U}^t\)}
        \STATE Construct candidate set \(\mathcal{C}_u^t\).
        \STATE Predict \(p_{u,a}^t\) for all \(a\in\mathcal{C}_u^t\) using~\eqref{eq:qos_risk_prediction}.
        \STATE Select \(a_u^{\star,t}\) by \eqref{eq:best_candidate}
        \STATE Trigger handover only if \eqref{eq:handover_condition} is satisfied
    \ENDFOR
\ENDFOR
\end{algorithmic}
\end{algorithm}

\subsection{Handover-Aware Access Selection}
\label{subsec:access_decision}

After training, the global model is deployed at the edge controller or at local access controllers. 
For a candidate $a \in \mathcal{C}_u^t$, the model estimates the probability of QoS failure $p_{u, a}^t$.  The access decision minimizes a handover-aware risk score:
\begin{equation}
    J_{u,a}^t
    =
    p_{u,a}^t
    +
    \eta_{\rm h}
    \mathbf{1}_{a\neq a_u^{t-1}},
    \label{eq:access_score}
\end{equation}
where \(\eta_{\rm h}\geq 0\) is the handover cost coefficient. The indicator $\mathbf{1}\left[a\neq a_u^{t-1}\right]$ equals one if selecting candidate $a$ triggers a handover, and zero otherwise. The best candidate is
\begin{equation}
    a_u^{\star,t}
    =
    \arg\min_{a\in\mathcal{C}_u^t}
    J_{u,a}^t.
    \label{eq:best_candidate}
\end{equation}
To avoid ping-pong handovers, the controller triggers a handover only if the candidate improves the current serving score by at least a hysteresis margin \(\xi_{\rm h}\) and the user has remained with the current serving node for at least \(T_{\rm dwell}\) slots:
\begin{equation}
    J_{u,a_u^{t-1}}^t
    -
    J_{u,a_u^{\star,t}}^t
    >
    \xi_{\rm h},
    \qquad
    t-t_u^{\rm last}
    \geq
    T_{\rm dwell},
    \label{eq:handover_condition}
\end{equation}
where \(t_u^{\rm last}\) is the most recent handover time of user \(u\). If \eqref{eq:handover_condition} is not satisfied, the user remains connected to \(a_u^{t-1}\). Finally, Algorithm~\ref{alg:fedqos} summarizes the proposed FedQoS approach.

\section{Experimental Setup}
\label{sec:experimental_setup}

We use Sionna RT and Sionna SYS~\cite{hoydis2022sionna,
hoydis2023sionnart} to generate physics-based network logs for
heterogeneous indoor-outdoor environments. The logs are partitioned by
observing access node, and each node trains only on its local data while
exchanging model parameters with the edge controller. The resulting
models are evaluated as both QoS-failure predictors and access-selection
policies.


\subsection{Sionna-Based Dataset Generation}
\label{subsec:sionna_dataset}

We construct a $200\,\mathrm{m}\times200\,\mathrm{m}$ campus containing
three multi-floor buildings, indoor APs, one outdoor BS, two quasi-static
UAV-mounted access nodes, outdoor walkways, and an event area. Indoor
structures, material-dependent walls, furniture, vegetation, vehicles,
and other outdoor obstacles create blockage and non-line-of-sight
conditions. The APs represent private-5G indoor small cells, the BS
represents an outdoor campus cell, and the UAVs remain at fixed hovering
positions during each simulation episode. Table~\ref{tab:exp_parameters}
summarizes the common simulation and learning parameters.

For each scenario, Sionna RT computes geometry-aware propagation between users and access nodes, while Sionna SYS generates QoS-related quantities such as received power, SINR,
BLER, throughput, access-node load, and resource utilization. The simulated users follow short indoor-outdoor trajectories, including normal occupancy, class transition, outdoor event, stress-event, and access-node degradation scenarios. These scenarios capture room/corridor usage, mobility between indoor and outdoor areas, local crowding, heavy-load conditions, and temporary radio or load degradation.

Each log entry is a user-candidate tuple $(u,a,t)$ from
$\mathcal{C}_u^t$, which contains the current serving node and the
top-$K$ feasible measurement candidates. Sionna probes these candidates
over the prediction window to obtain future throughput and BLER and
derive the QoS-failure labels. In practice, serving-link outcomes are
logged directly, while unselected links require probing, dual
connectivity, or controlled exploration. Samples remain at their
observing access nodes and are locally split into 70\%/15\%/15\%
training, validation, and test sets.

We evaluate two naturally non-IID conditions using the same campus,
radio configuration, user-density ranges, prediction horizon, and
learning settings. The \emph{mild non-IID} condition mainly contains
normal and class-transition traffic, uses service rates of
$1/5/10$~Mbps, and rarely includes AP outages. The
\emph{severe non-IID} condition uses $2/8/15$~Mbps, increases
event-, stress-, and outage-related traffic, and introduces higher
client heterogeneity through stronger room skew, user churn, probing,
and reduced candidate sets.

\begin{table}[!ht]
\vspace*{5pt}
\centering
\caption{Default simulation and learning parameters.}
\label{tab:exp_parameters}
\begin{tabular}{l|c}
\hline
\textbf{Parameter} & \textbf{Value} \\
\hline
Campus area & $200\,\mathrm{m} \times 200\,\mathrm{m}$ \\ 
Buildings & 3 multi-floor buildings (A, B, C) \\ 
Floors / APs per building & A: 2/4, B: 3/9, C: 4/16 \\ 
Floor height & $4\,\mathrm{m}$ \\ 
Access nodes & 29 APs, 1 outdoor BS, 2 UAVs \\ 
Carrier frequency & $3.5\,\mathrm{GHz}$ \\ 
Access-node transmit powers & AP/BS/UAV: 24/37/30 dBm \\ 
Access-node antenna gains & AP/BS/UAV: 4/12/7 dBi \\ 
OFDM subcarriers / spacing & 1596 / 30 kHz \\ 
Simulated bandwidth & 47.88 MHz \\ 
OFDM symbols per slot & 14 \\ 
BLER target $\epsilon_{\max}$ & 0.10 \\
Decision-slot duration & 1 second \\
Prediction window & 5 seconds \\
Candidate construction & Top-$K$ measurement candidates \\
Top-$K$ candidates for mild setting & 4 \\
Top-$K$ candidates for severe setting & 3 \\
AP-outage scenario share & Mild: $5\%$; Severe: $30\%$ \\
Handover factors $\eta_h$, $\xi_h$, $T_{dwell}$ & 0.08, 0.04, 3 time slots \\
\hline
ML model & Multi-Layer Perceptron \\
Hidden layer & 2 layers with size (96, 96) \\
Activations & ReLU, Sigmoid (output layer)\\
Federated rounds & 50 \\
Local epochs and batch size & 1, 128 \\
Optimizer and learning rate & Adam with lr=0.001 \\
FedProx/FedQoS coefficient ($\mu$) & 0.01 \\
FedQoS parameters $(q,\lambda)$ & $(0.75, 0.5)$ \\
\hline 
\end{tabular}
\end{table}

\subsection{Baselines and Metrics}
\label{subsec:baselines_metrics}

The proposed \textbf{FedQoS} algorithm is compared with four learning
baselines. \textbf{Centralized} pools all training data at a central
server and serves as a centralized learning reference.
\textbf{Local-only} trains an independent model at each access node
without model aggregation. \textbf{FedAvg} performs standard
sample-size-weighted federated averaging, whereas \textbf{FedProx} adds
a proximal regularization term during local training to mitigate client
drift under heterogeneous data.

We also compare FedQoS with several non-learning access-selection
policies. \textbf{Current serving} keeps each user connected to its
existing serving node. \textbf{Strongest SINR} selects the candidate
with the highest instantaneous SINR. \textbf{Load-aware RSRP} jointly
considers received signal power and access-node load. \textbf{AP-first}
prioritizes AP candidates whenever available and falls back to a BS or
UAV otherwise. \textbf{Historical QoS} selects the candidate with the
lowest historical QoS-failure rate in the training logs. Finally,
\textbf{Oracle} selects candidates using their realized future QoS
outcomes and is included only as a non-implementable performance
reference.

For QoS-risk prediction, we report balanced accuracy, recall, and F1-score. Balanced accuracy accounts for the imbalance between successful and failed QoS samples, recall measures the ability to detect imminent QoS failures, and the F1-score summarizes the balance between failure detection and false alarms. Higher values indicate better prediction performance.

For access-selection performance, we report the
QoS-failure rate, mean throughput, and mean BLER of the selected links. The QoS-failure rate is the
primary system-level metric because the objective is to avoid future QoS
violations. A lower failure rate and BLER indicate more reliable access
selection, whereas a higher throughput indicates better data-rate
performance. 

\section{Results and Discussion}
\label{sec:results}

We report results under the mild and severe non-IID conditions described in Section~\ref{subsec:sionna_dataset}. Learning-based results are averaged over ten random seeds and reported as mean $\pm$ standard deviation.

\begin{table*}[t]
\vspace*{6pt}
\centering
\caption{Prediction and access-selection performance over ten seeds.
Boldface marks the best mean among Local-only, FedAvg, FedProx, and FedQoS, with standard deviation breaking ties at the displayed precision.}
\label{tab:main_results}
\scriptsize
\setlength{\tabcolsep}{3.2pt}
\renewcommand{\arraystretch}{0.98}

\textbf{(a) QoS-failure prediction}\par\vspace{0.3mm}
\begin{tabular}{lccc|ccc}
\toprule
& \multicolumn{3}{c|}{Mild non-IID}
& \multicolumn{3}{c}{Severe non-IID} \\
Method
& BAcc. $\uparrow$
& Recall $\uparrow$
& F1 $\uparrow$
& BAcc. $\uparrow$
& Recall $\uparrow$
& F1 $\uparrow$ \\
\midrule
Centralized
& $0.9493 \pm 0.0044$
& $0.9174 \pm 0.0100$
& $0.8902 \pm 0.0027$
& $0.9688 \pm 0.0026$
& $0.9617 \pm 0.0071$
& $0.9475 \pm 0.0021$ \\ \hline

Local-only
& $0.9234 \pm 0.0041$
& $0.8738 \pm 0.0089$
& $0.8420 \pm 0.0040$
& $0.9466 \pm 0.0014$
& $0.9344 \pm 0.0028$
& $0.9114 \pm 0.0019$ \\

FedAvg
& $0.9421 \pm 0.0039$
& $0.9089 \pm 0.0102$
& $\mathbf{0.8675 \pm 0.0023}$
& $0.9670 \pm 0.0022$
& $0.9599 \pm 0.0077$
& $0.9441 \pm 0.0012$ \\

FedProx
& $0.9407 \pm 0.0036$
& $0.9056 \pm 0.0088$
& $0.8675 \pm 0.0030$
& $0.9671 \pm 0.0024$
& $0.9606 \pm 0.0078$
& $0.9440 \pm 0.0012$ \\

FedQoS
& $\mathbf{0.9440 \pm 0.0049}$
& $\mathbf{0.9137 \pm 0.0125}$
& $0.8672 \pm 0.0025$
& $\mathbf{0.9679 \pm 0.0029}$
& $\mathbf{0.9617 \pm 0.0085}$
& $\mathbf{0.9451 \pm 0.0014}$ \\
\bottomrule
\end{tabular}

\vspace{0.7mm}
\textbf{(b) Access-selection performance}\par\vspace{0.3mm}
\begin{tabular}{lccc|ccc}
\toprule
& \multicolumn{3}{c|}{Mild non-IID}
& \multicolumn{3}{c}{Severe non-IID} \\
Method
& Failure $\downarrow$
& Throughput (Mbps) $\uparrow$
& BLER $\downarrow$
& Failure $\downarrow$
& Throughput (Mbps) $\uparrow$
& BLER $\downarrow$ \\
\midrule
Current serving
& $0.1083$ & $20.0043$ & $0.0459$
& $0.2969$ & $18.6359$ & $0.0418$ \\

Strongest SINR
& $0.0962$ & $20.3100$ & $0.0386$
& $0.2795$ & $19.6464$ & $0.0413$ \\

Load-aware RSRP
& $0.0928$ & $19.6950$ & $0.0374$
& $0.2762$ & $20.6446$ & $0.0414$ \\

AP-first
& $0.0983$ & $20.5194$ & $0.0395$
& $0.2706$ & $20.7833$ & $0.0418$ \\

Historical QoS
& $0.0905$ & $20.4140$ & $0.0363$
& $0.2668$ & $21.3871$ & $0.0365$ \\
\midrule
Centralized
& $0.0025 \pm 0.0005$
& $18.0545 \pm 0.0744$
& $0.0007 \pm 0.0002$
& $0.1357 \pm 0.0007$
& $19.3097 \pm 0.0669$
& $0.0091 \pm 0.0003$ \\

Local-only
& $0.0055 \pm 0.0004$
& $17.6719 \pm 0.0503$
& $0.0020 \pm 0.0002$
& $0.1441 \pm 0.0009$
& $19.1336 \pm 0.0316$
& $0.0105 \pm 0.0003$ \\

FedAvg
& $0.0036 \pm 0.0005$
& $\mathbf{17.7452 \pm 0.0443}$
& $0.0012 \pm 0.0002$
& $0.1356 \pm 0.0008$
& $\mathbf{19.2150 \pm 0.0321}$
& $0.0095 \pm 0.0004$ \\

FedProx
& $0.0033 \pm 0.0003$
& $17.7062 \pm 0.0387$
& $0.0010 \pm 0.0001$
& $0.1358 \pm 0.0003$
& $19.1920 \pm 0.0245$
& $\mathbf{0.0095 \pm 0.0002}$ \\

FedQoS
& $\mathbf{0.0031 \pm 0.0002}$
& $17.6586 \pm 0.0476$
& $\mathbf{0.0009 \pm 0.0001}$
& $\mathbf{0.1356 \pm 0.0003}$
& $19.1817 \pm 0.0346$
& $0.0096 \pm 0.0002$ \\
\midrule
Oracle
& $0.0004$ & $27.1080$ & $0.0004$
& $0.1289$ & $25.0075$ & $0.0041$ \\
\bottomrule
\end{tabular}
\end{table*}

\subsection{Prediction and Access-Selection Performance}
\label{subsec:main_results}

Table~\ref{tab:main_results}(a) compares the QoS-failure prediction
performance. In both settings, all federated methods clearly outperform
local-only training, demonstrating that collaboration across access-node
clients compensates for the limited and heterogeneous observations
available at individual nodes. As expected, the centralized model
generally provides the strongest overall reference because it has direct
access to the complete training dataset.

In the mild non-IID setting, FedQoS achieves the highest balanced accuracy and recall among the federated methods. Its F1-score remains nearly identical to those of FedAvg and FedProx, indicating that the improved failure sensitivity does not materially degrade the overall classification balance. In particular, FedQoS favors the detection of imminent QoS violations, which is desirable because missed failures can lead directly to poor access-selection decisions. FedProx and FedAvg provide slightly higher displayed F1-score, but the differences are small relative to the variability across random seeds.

Under severe non-IID data, FedQoS achieves the highest mean balanced
accuracy, recall, and F1 among the federated methods. Although the numerical differences from FedAvg and FedProx remain modest, FedQoS consistently preserves failure-detection performance under stronger topology-induced heterogeneity, increased user density, and partial AP unavailability. The substantial gap between all federated models and local-only training also confirms that isolated access nodes cannot learn sufficiently general QoS-failure predictors from their own observations.

Overall, FedQoS maintains prediction quality comparable to conventional
federated learning under mild non-IID data and provides the strongest
failure-oriented predictive performance under the severe non-IID
condition.

Table~\ref{tab:main_results}(b) evaluates the downstream access-selection
performance. The conventional policies represent different practical
selection principles, including maintaining the current serving node,
selecting the strongest SINR, considering both RSRP and load, prioritizing
AP connectivity, and exploiting historical QoS. However, these policies
rely mainly on instantaneous or past measurements and cannot directly
anticipate future QoS violations. Consequently, their failure rates and
BLER values remain substantially higher than those of the learning-based policies in both settings.

In the mild non-IID setting, FedQoS achieves the lowest mean QoS-failure rate and BLER among the federated methods. Its failure rate is reduced by approximately $13.9\%$ relative to FedAvg and $6.1\%$ relative to FedProx. The reductions relative to local-only training and historical QoS are approximately $43.6\%$ and $96.6\%$, respectively. 
These gains show that the overall FedQoS design improves not only
failure prediction but, more importantly, the access decisions derived
from the predicted probabilities.

FedQoS obtains these reliability improvements with only a small
throughput reduction. Its average throughput is approximately $0.49\%$
below FedAvg and $0.27\%$ below FedProx. This behavior is consistent
with the objective of FedQoS: rather than selecting links solely for
their instantaneous data rate, the policy favors candidates with a lower predicted risk of future QoS violation. The resulting reduction in failure rate and BLER therefore outweighs the marginal throughput cost.

The severe non-IID setting is substantially more challenging, as indicated by the higher failure rates of all policies. Nevertheless, all learning-based approaches still markedly outperform the conventional heuristics. FedQoS ties FedAvg at the displayed precision for the lowest federated failure rate and remains close to FedAvg and FedProx in both throughput and BLER. Its throughput is less than $0.2\%$ below FedAvg, while the BLER difference is only $10^{-4}$. Hence, the three federated
methods operate at very similar system-level performance in this
setting, with FedQoS retaining a reliability-oriented operating point
under stronger client heterogeneity.

The oracle achieves the lowest failure rate and BLER and the highest
throughput because it selects access links using future QoS outcomes.
It is therefore included only as an unattainable reference rather than
an implementable competitor. 
Overall, FedQoS provides the clearest reliability gains under mild non-IID data and remains robust and competitive under the more difficult severe non-IID condition.

\subsection{Sensitivity to $q$ and $\lambda$}
\label{subsec:sensitivity}

\begin{figure*}[t]
    \centering
    \begin{subfigure}{0.45\textwidth}
        \centering
        \includegraphics[width=\linewidth]
        {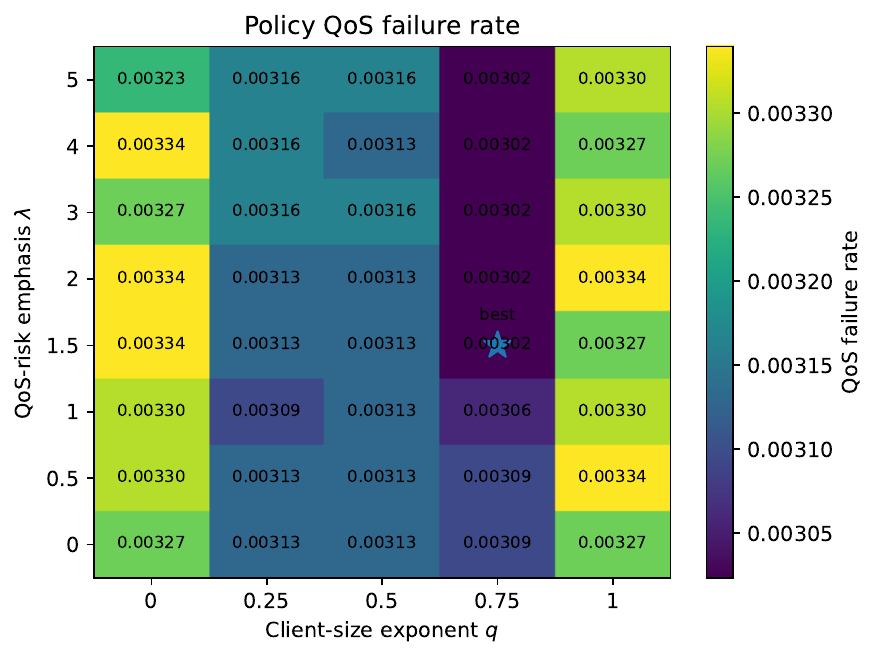}
        \caption{Mild non-IID setting.}
    \end{subfigure}
    \hfill
    \begin{subfigure}{0.45\textwidth}
        \centering
        \includegraphics[width=\linewidth]
        {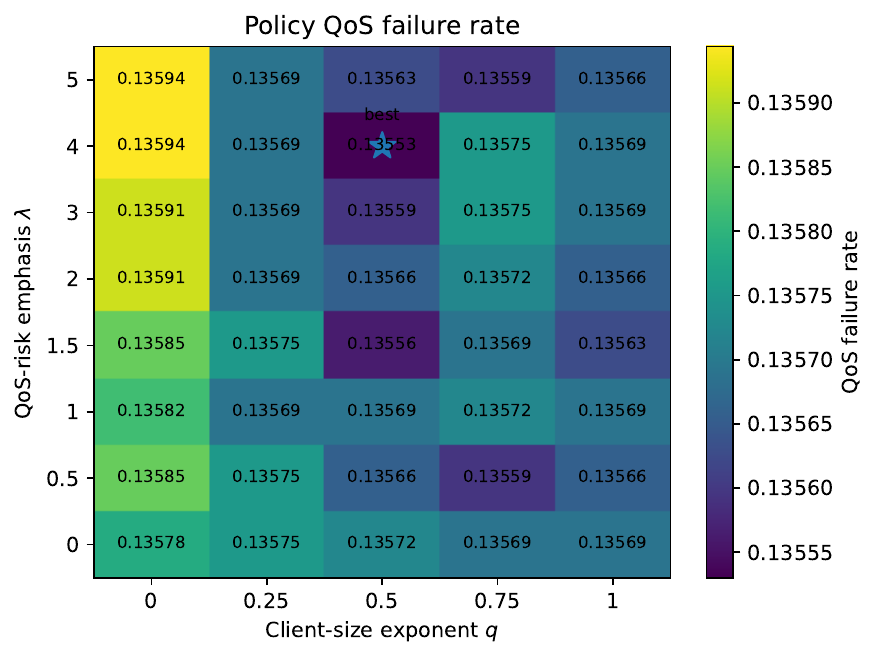}
        \caption{Severe non-IID setting.}
    \end{subfigure}
    \caption{Sensitivity of the policy QoS-failure rate to the
    client-size exponent $q$ and QoS-risk emphasis parameter
    $\lambda$.}
    \label{fig:sensitivity}
\end{figure*}

We evaluate the sensitivity of FedQoS to the client-size exponent $q$
and the QoS-risk emphasis parameter $\lambda$. Since the policy
QoS-failure rate is the primary system-level objective,
Fig.~\ref{fig:sensitivity} reports its variation over the evaluated
parameter grid.

In the mild non-IID setting, the QoS-failure rate varies only from approximately
$0.00302$ to $0.00334$. The lowest value is obtained near
$(q,\lambda)=(0.75,1.5)$, while a broad region with moderate or large
$q$ provides nearly identical performance. The common setting used in
the main experiments, $(q,\lambda)=(0.75,0.5)$, also lies within this
low-failure region.

The severe non-IID setting is similarly stable, with the QoS-failure rate
remaining within approximately $0.13553$--$0.13594$ across the complete
grid. Although the minimum is observed near
$(q,\lambda)=(0.5,4)$, several neighboring combinations yield almost
indistinguishable results. This indicates a broad performance plateau
rather than a sharply localized optimum.

Across both settings, $q$ has a more visible influence than $\lambda$,
particularly when $q$ is close to zero. Hence, controlling the
contribution of clients with highly unequal dataset sizes is more
important than aggressively increasing the risk emphasis. Once a
reasonable client-size weighting is selected, $\lambda$ mainly
fine-tunes the resulting policy.

Overall, the sensitivity analysis shows that FedQoS remains stable over
a broad range of parameter values and that the reported policy gains do
not depend on narrowly tuned values of $q$ and $\lambda$.

\section{Conclusion}
\label{sec:conclusion}

This paper presented FedQoS, a federated QoS-risk learning framework for reliable access selection in heterogeneous indoor-outdoor environments.
Instead of relying only on instantaneous signal strength, FedQoS learns
the future QoS-failure risk of candidate access links from distributed
network observations collected at APs, UAVs, and the BS. The learned risk scores are then used to support reliability-aware access decisions without centralizing user-level data. Simulation results on physics-based synthetic campus datasets show that learning-based access selection substantially reduces QoS failures and BLER compared with conventional signal-based and historical-QoS heuristics. FedQoS achieves near-centralized prediction performance and remains effective under both mild and severe non-IID conditions. Its stable performance across a broad range of aggregation parameters further demonstrates its potential for reliable access selection in dynamic heterogeneous networks.

\ifCLASSOPTIONcaptionsoff
  \newpage
\fi

\section*{Acknowledgment}
\small{
This work was supported in part by the Luxembourg National Research
Fund (FNR) through the CHIST-ERA SHIELD project under Grant
INTER/CHIST24/19023763/SHIELD, and in part by the FNR AFR program
through the FULFIL project under Grant 2000141.}

\bibliographystyle{IEEEtran}
\bibliography{IEEEabrv,ref_conf.bib} 

@article{ye2013user,
  title={User association for load balancing in heterogeneous cellular networks},
  author={Ye, Qiaoyang and Rong, Beiyu and Chen, Yudong and Al-Shalash, Mazin and Caramanis, Constantine and Andrews, Jeffrey G},
  journal={IEEE Transactions on Wireless Communications},
  volume={12},
  number={6},
  pages={2706--2716},
  year={2013},
  publisher={IEEE}
}

@article{liu2016user,
  title={User association in 5G networks: A survey and an outlook},
  author={Liu, Dantong and Wang, Lifeng and Chen, Yue and Elkashlan, Maged and Wong, Kai-Kit and Schober, Robert and Hanzo, Lajos},
  journal={IEEE Communications Surveys \& Tutorials},
  volume={18},
  number={2},
  pages={1018--1044},
  year={2016},
  publisher={IEEE}
}

@article{kim2023distributed,
  title={Distributed resource allocation and user association for max-min fairness in HetNets},
  author={Kim, Yeongjun and Jang, Jonggyu and Yang, Hyun Jong},
  journal={IEEE Transactions on Vehicular Technology},
  volume={73},
  number={2},
  pages={2983--2988},
  year={2023},
  publisher={IEEE}
}

@inproceedings{zhou2014qos,
  title={QoS-aware user association for load balancing in heterogeneous cellular networks},
  author={Zhou, Tianqing and Huang, Yongming and Huang, Wei and Li, Shidang and Sun, Yuan and Yang, Luxi},
  booktitle={2014 IEEE 80th Vehicular Technology Conference (VTC2014-Fall)},
  pages={1--5},
  year={2014},
  organization={IEEE}
}

@article{kong2017delay,
  title={Delay-optimal biased user association in heterogeneous networks},
  author={Kong, Fancheng and Sun, Xinghua and Leung, Victor CM and Zhu, Hongbo},
  journal={IEEE Transactions on Vehicular Technology},
  volume={66},
  number={8},
  pages={7360--7371},
  year={2017},
  publisher={IEEE}
}

@article{cao2021user,
  title={User access control in open radio access networks: A federated deep reinforcement learning approach},
  author={Cao, Yang and Lien, Shao-Yu and Liang, Ying-Chang and Chen, Kwang-Cheng and Shen, Xuemin},
  journal={IEEE Transactions on Wireless Communications},
  volume={21},
  number={6},
  pages={3721--3736},
  year={2021},
  publisher={IEEE}
}

@inproceedings{tehrani2021federated,
  title={Federated deep reinforcement learning for the distributed control of NextG wireless networks},
  author={Tehrani, Peyman and Restuccia, Francesco and Levorato, Marco},
  booktitle={2021 IEEE international symposium on dynamic spectrum access networks (DySPAN)},
  pages={248--253},
  year={2021},
  organization={IEEE}
}

@article{zhao2019deep,
  title={Deep reinforcement learning for user association and resource allocation in heterogeneous cellular networks},
  author={Zhao, Nan and Liang, Ying-Chang and Niyato, Dusit and Pei, Yiyang and Wu, Minghu and Jiang, Yunhao},
  journal={IEEE Transactions on Wireless Communications},
  volume={18},
  number={11},
  pages={5141--5152},
  year={2019},
  publisher={IEEE}
}

@article{lim2020federated,
  title={Federated learning in mobile edge networks: A comprehensive survey},
  author={Lim, Wei Yang Bryan and Luong, Nguyen Cong and Hoang, Dinh Thai and Jiao, Yutao and Liang, Ying-Chang and Yang, Qiang and Niyato, Dusit and Miao, Chunyan},
  journal={IEEE communications surveys \& tutorials},
  volume={22},
  number={3},
  pages={2031--2063},
  year={2020},
  publisher={IEEE}
}

@article{bega2019deepcog,
  title={DeepCog: Optimizing resource provisioning in network slicing with AI-based capacity forecasting},
  author={Bega, Dario and Gramaglia, Marco and Fiore, Marco and Banchs, Albert and Costa-Perez, Xavier},
  journal={IEEE Journal on Selected Areas in Communications},
  volume={38},
  number={2},
  pages={361--376},
  year={2019},
  publisher={IEEE}
}

@article{mollel2021survey,
  title={A survey of machine learning applications to handover management in 5G and beyond},
  author={Mollel, Michael S and Abubakar, Attai Ibrahim and Ozturk, Metin and Kaijage, Shubi Felix and Kisangiri, Michael and Hussain, Sajjad and Imran, Muhammad Ali and Abbasi, Qammer H},
  journal={IEEE Access},
  volume={9},
  pages={45770--45802},
  year={2021},
  publisher={IEEE}
}

@article{li2020federated,
  title={Federated optimization in heterogeneous networks},
  author={Li, Tian and Sahu, Anit Kumar and Zaheer, Manzil and Sanjabi, Maziar and Talwalkar, Ameet and Smith, Virginia},
  journal={Proceedings of Machine learning and systems},
  volume={2},
  pages={429--450},
  year={2020}
}

@article{hoydis2022sionna,
  author  = {Hoydis, Jakob and Cammerer, Sebastian and {A\"it Aoudia}, Fay\c{c}al
             and Vem, Avinash and Binder, Nikolaus and Marcus, Guillermo and Keller, Alexander},
  title   = {{Sionna}: An Open-Source Library for Next-Generation Physical Layer Research},
  journal = {arXiv preprint arXiv:2203.11854},
  year    = {2022}
}

@article{hoydis2023sionnart,
  author  = {Hoydis, Jakob and {A\"it Aoudia}, Fay\c{c}al and Cammerer, Sebastian
             and Euchner, Florian and Nimier-David, Merlin and ten Brink, Stephan and Keller, Alexander},
  title   = {{Sionna RT}: Differentiable Ray Tracing for Radio Propagation Modeling},
  journal = {arXiv preprint arXiv:2303.11103},
  year    = {2023}
}

%







\end{document}